\documentclass[10pt,twocolumn,letterpaper]{article}
\usepackage{cvpr}
\usepackage{times}
\usepackage{epsfig}
\usepackage{graphicx}
\usepackage{amsmath}
\usepackage{amssymb}
\usepackage{multirow}

\usepackage[breaklinks=true,bookmarks=false]{hyperref}

\cvprfinalcopy 

\def\cvprPaperID{****} 

\begin{document}

\title{Dense-Captioning Events in Videos: \\SYSU Submission to ActivityNet Challenge 2020}

\iftrue
\author{Teng Wang, Huicheng Zheng, Mingjing Yu\\
School of Data and Computer Science, Sun Yat-sen University, China\\
Key Laboratory of Machine Intelligence and Advanced Computing, Ministry of Education, China\\
Guangdong Province Key Laboratory of Information Security Technology, China\\
{\tt\small {zhenghch@mail}.sysu.edu.cn}\\
}
\fi
\maketitle

\begin{figure*}
	\centering
	\includegraphics[width=1.0\linewidth]{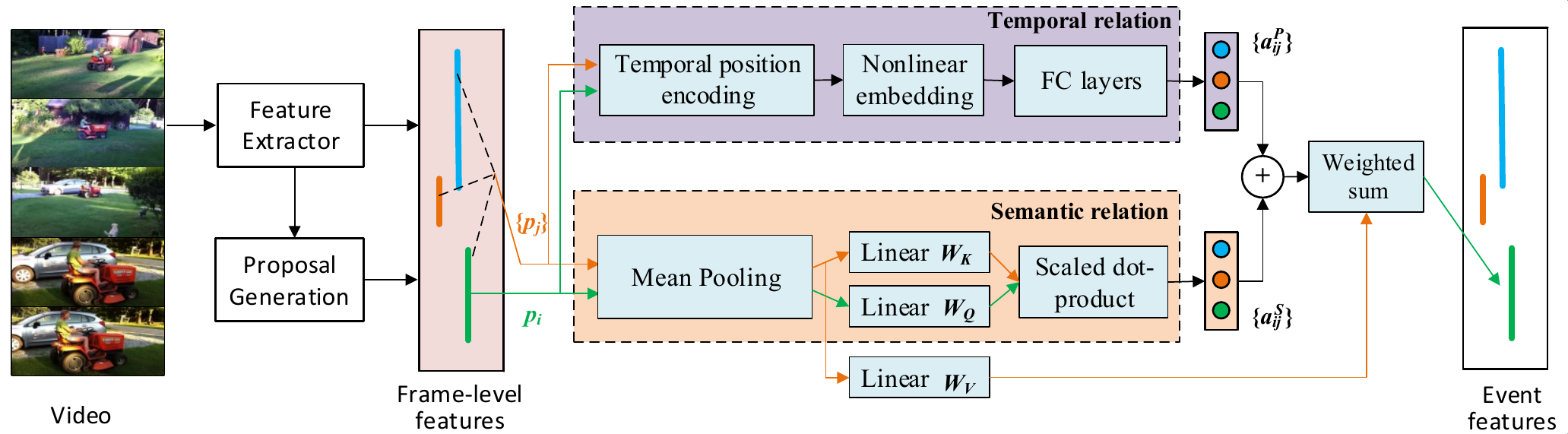}
	\caption{The proposed encoder.}
	\label{fig:encoder}
\end{figure*}

\begin{abstract}
   This technical report presents a brief description of our submission to the dense video captioning task of ActivityNet Challenge 2020. Our approach follows a two-stage pipeline: first, we extract a set of temporal event proposals; then we propose a multi-event captioning model to capture the event-level temporal relationships and effectively fuse the multi-modal information. Our approach achieves a 9.28 METEOR score on the test set. Code is available at \url{https://github.com/ttengwang/dense-video-captioning-pytorch}.
\end{abstract}

\section{Introduction}
Dense video captioning attracts increasing attention in recent years, whose goal is to localize and describe all events in an untrimmed video. Different from traditional video captioning which only described a single event, dense video captioning requires a comprehensive understanding of the long-term temporal structure and the semantic relationships among a sequence of events. Previous methods~\cite{chen2019activitynet,krishna2017dense,wang2018bidirectional} have employed different types of contexts for constructing event representation, e.g., neighboring regions within an expanded receptive field~\cite{chen2019activitynet}, event-level semantic  attention~\cite{krishna2017dense}, and clip-level recurrent features~\cite{wang2018bidirectional,chen2019activitynet,yang2018hierarchical}. Although promising progress has been achieved, their context modeling missed the perception of the temporal structure of the event sequence, i.e. the temporal positions and lengths of other events. As a consequence, the temporal relationship between events is not fully exploited in the captioning stage. In this work, we propose to explore the temporal relation between events in the encoding phase. Furthermore, we also design a cross-modal gating (CMG) block for hierarchical RNN, which can adaptively estimate the weight of linguistic information and visual information for better caption generation.

\section{Method}
The overall framework contains three parts, i.e., the feature extractor, the temporal event proposal model, and the event captioning model.

\subsection{Feature Extractor}
We divide the video into several non-overlapping clips with a stride of 0.5s and extract the frame-level features by a TSN~\cite{xiong2016cuhk} pretrained on ActivityNet datasets for the action recognition task. We concatenate the feature vectors in optical flow modality and RGB modality to construct the frame-level representation,  which is utilized for the temporal event proposal module (TEP) and the event captioning module (EC).

\subsection{Temporal Event Proposal}

Accurate event proposals generation is the basis for further captioning. For TEP, we adopt an off-the-shell DBG~\cite{lin2019fast} to detect the top 100 proposals for each video. Since the number of proposals in the ground-truth annotation is usually small (around 3.7 per video) in average, we follow Chen et al.~\cite{chen2019activitynet} to perform a modified event sequence selection network (ESGN)~\cite{Mun2019stream} to predict a subset of candidate proposals. The selection process can achieve a good balance between precision and recall, especially when the number of proposals is relatively small. After selection, the number of output proposals per video is around 2.4 on average. The average precision and recall on the validation set across tIOU$\in\{0.3, 0.5, 0.7, 0.9\}$ is 66.63\% and 40.09\%, respectively.

\begin{figure*}
	\centering
	\includegraphics[width=0.75\linewidth]{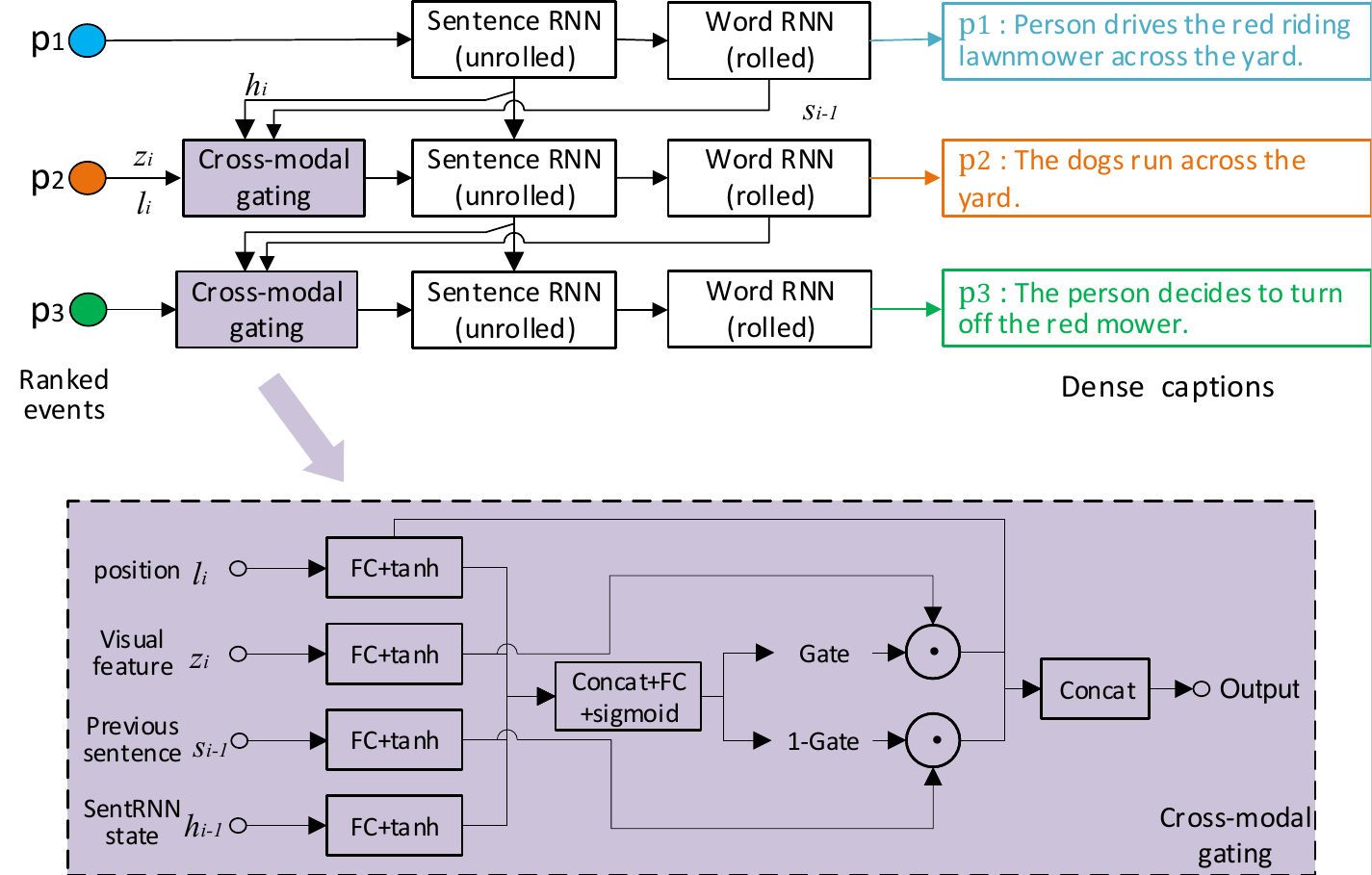}
	\caption{The proposed Decoder.}
	\label{fig:decoder}
\end{figure*}

\subsection{Event Captioning}
The event captioning model follows an encoder-decoder architecture. For the visual encoder, we follow Wang et al.~\cite{wang2020} to adopt the temporal-semantic relation module (TSRM) to capture rich relationships between events in terms of both temporal structure and semantic meaning.  For the language decoder, we develop a gated hierarchical RNN, aided by a cross-modal gate to strike a balance of visual information and linguistic information when captioning the next sentence.  The encoder and the decoder are illustrated in Fig.~\ref{fig:encoder} and Fig.~\ref{fig:decoder}, respectively.

{\textbf{Visual Encoder.}} TSRM contains two branches, i.e., a temporal relation branch and a semantic relation branch. Each branch estimates the relation score between the target proposal $p_i$ and the other proposals $\{p_j\}$, and then we fuse the two types of scores by addition. The relational features of an event are obtained by weighted summation of features of all proposals conditioned on the relation scores.

For the temporal relation branch, we encode the relative length and the relative distance between each pair of proposals, and then put the position encoding into a non-linear embedding~\cite{vaswani2017attention} and an FC layers with an output size of 1. For the semantic relation branch, we first obtain the proposals' initial representation by the mean pooling of the frame-level features, then we adopt scaled dot-product attention~\cite{vaswani2017attention} to calculate the semantic similarity between proposals.

\begin{table*}
	\begin{center}
		\begin{tabular}{ccc|ccc}
			\hline
			TSRM & CMG & sent RNN & \textbf{METEOR} & BLEU@4 & CIDEr\\
			\hline
			$\times$ & $\times$ & $\times$ & 9.96 & 1.67 &32.67\\
			$\surd$  & $\times$ &$\times$ &11.16 & 2.73 & 49.15 \\
			
			$\times$ & $\times$ & $\surd$ & 10.76 & 2.28 & 40.51\\
			$\times$ & $\surd$ & $\surd$ & 11.31 &2.74 &48.37\\
			$\surd$  & $\surd$ & $\surd$ & 11.49 &2.85 &49.34\\
			\hline
		\end{tabular}
	\end{center}
	\caption{Ablation study for event captioning model on validation set with ground-truth proposals. Note that the validation set has two independent annotations, we only use the ground-truth proposal in val\_1 as the input proposals. The performance is evaluated on both annotations according to the official evaluation code.}
	\label{table:ablation}
\end{table*}

{\textbf{Language Decoder.}}
The function of decoder is to translate the visual representation produced by the encoder into target modality. Different from models in traditional image/video captioning, the target output of the decoder in dense video captioning is a set of sentences instead of one. To increase the coherence among sentences, hierarchical RNN (a sentence RNN plus a word RNN) based deocder is widely used for multi-sentence captioning~\cite{yu2016video,Mun2019stream,chen2019activitynet}. The sentence RNN stores multi-modal information of all previous events and guides the generation of the next sentence. To enhance the multi-modal fusion, we propose a cross-modal gating (CMG) block to adaptively balance the visual and linguistic information. Specifically, the inputs of cross-modal gating block are four folds: 1) the position embedding $l_i$ of proposal $p_i$, 2) the proposal's feature vector $z_i$, which is the concatenation of the output of TSRM module and the mean pooling of the frame-level features within $p_i$, 3) the last hidden state $s_{i-1}$ in the word RNN of the previous sentence, and 4) the previous hidden state $h_{i-1}$ of the sentence RNN. Motivated by Wang et al.~\cite{wang2018bidirectional}, we use gating mechanism to balance the linguistic information $s_{i-1}$ and the visual information $z_i$. The gate $g_i$ is calculated by an FC layer with sigmoid activation function. We use $g_t$ and $1-g_t$ to gate the information of visual features and linguistic features, respectively. The word RNN is implemented as an attention-enhanced RNN, which adaptively select the salient frames within the proposal $p_i$ for word prediction.

\begin{table}
	
	\renewcommand\arraystretch{1.2}
	\newcommand{\tabincell}[2]{\begin{tabular}{@{}#1@{}}#2\end{tabular}}
	\begin{center}
		\setlength{\tabcolsep}{0.4mm}{
			\begin{tabular}{c|ccc}
				\hline
				 & \multicolumn{2}{c}{Validation set} & Test set\\
				 & GT prop. & Learnt prop. & Learnt prop. \\
				\hline
				Cross-entropy & 11.49  & 7.65 & - \\
				+SCST & 14.18(+2.69) & 9.71(+2.06) & -\\
				+ Larger train. set & 14.64$^*$(+0.46) & 10.26$^*$(+0.55) & 9.17 \\
				+ ensemble & 14.85$^*$(+0.21) & 10.31$^*$(+0.05) & 9.28(+0.11) \\
				\hline
			\end{tabular}
		}
	\end{center}
	\caption{Performance of different training scheme and model ensemble on validation set. $^*$ incicates evaluation results on the small validation set. }
	\label{table:tricks}
\end{table}

\section{Experiments}
\subsection{Experimental Setting}

\textbf{Inplementation\ Details.} For data processing, we build a vocabulary that only takes into consideration those words that occurred at least 5 times. Sentences longer than 30 words have been truncated. For the event captioning model, we adopt LSTM as the RNN in the decoder. The hidden units of LSTMs and all FC layers are set to be 512.

We first train the event captioning module based on ground-truth proposals using cross-entropy loss. Afterwards, to address the exposure bias problem and boost the performance, we continually train the model by self-critical sequence training (SCST)~\cite{Rennie2017self} based on learnt event sequences. For each video, we sample 24 event sequences from the output results of DBG~\cite{lin2019fast} for SCST. The reward is set to be the METEOR score.

\textbf{Dataset.}
We use the ActivityNet Captions dataset to evaluate the performance of our method. We follow the official split, which assigns 10,009/4,917/5,044 videos for training/validation/testing. In our final submission, we train the event captioning model using SCST with an enlarged training set. Specifically, we randomly select $\sim$80\% videos from the validation set and add them to the official training set. The modified split contains 13,926/1,000 videos for training/validation.

\textbf{Evaluation Metrics.} We use the official evaluation tool to measure the ability of our model in both localizing and captioning events. Specifically, the average precision is computed across tIoU thresholds of 0.3, 0.5, 0.7, and 0.9. The precision of generated captions is measured by traditional evaluation metrics in video captioning: BLEU, METEOR, and CIDEr.

\subsection{Performance Evaluation}

We show the ablation study for event captioning in Table~\ref{table:ablation}. The first row in the table shows that the lacking of event-event interaction leads to poor performance of generated captions. When the sentence RNN or TSRM is incorporated, the generated sentence has the perception of previous events, thus a significant performance improvement can be achieved. The proposed CMG brings the model a big advantage at effective multi-modal fusion, which further boosts the captioning capability of the hierarchical RNN.

We also investigate the performance of training schemes and model ensemble in Table~\ref{table:tricks}. When using SCST after the cross-entropy training, the METEOR score increases considerably from 11.49/7.65 to 14.18/9.71. When using the enlarged training set, the performance can obtain further improvement. Our single model achieves a 9.17 METEOR score on the challenging test set. In our final submission, we use an ensemble of three models with different seeds, which achieves a 9.28 METEOR score.

\subsection{Conclusion}

In this paper, we present a dense video captioning system with two plug-and-play modules, i.e. TSRM and CMG. TSRM aims to enhance the event-level representation by capturing rich relationships between events in terms of both temporal structure and semantic meaning. CMG is designed to effectively fuse the linguistic features and visual features in hierarchical RNN. Experimental results on ActivityNet Captions verify the effectiveness of our model.

\subsection{Acknowledgement}
This work was supported in part by the National Natural Science Foundation of China under Grant 61976231, Grant U1611461, Grant 61172141, Grant 61673402, and Grant 61573387.

{\small
\bibliographystyle{ieee_fullname}
\bibliography{Reference}
}

\end{document}